\pdfoutput=1

\documentclass[11pt]{article}

\usepackage{acl}

\usepackage{times}
\usepackage{latexsym}
\usepackage{inconsolata}
\usepackage{amsfonts}
\usepackage{soul}
\usepackage{colortbl}
\usepackage{multirow}
\usepackage{xspace}
\usepackage{amsmath}
\usepackage{graphicx}
\usepackage{booktabs}
\usepackage{algorithm}
\usepackage{algpseudocode}
\usepackage{bbm}

\usepackage[T1]{fontenc}

\usepackage[utf8]{inputenc}

\usepackage{microtype}

\usepackage{inconsolata}

\newcommand{\method}{EqualizeIR\xspace}

\title{\method: Mitigating Linguistic Biases in Retrieval Models}

\author{Jiali Cheng \quad Hadi Amiri \\
  University of Massachusetts Lowell \\
  \texttt{\{jiali\_cheng, hadi\_amiri\}@uml.edu} \\}

\begin{document}
\maketitle
\begin{abstract}
This study finds that existing information retrieval (IR) models show significant biases based on the linguistic complexity of input queries, performing well on linguistically simpler (or more complex) queries while underperforming on linguistically more complex (or simpler) queries.
To address this issue, we propose \method, a framework to mitigate linguistic biases in IR models. \method uses a {\em linguistically biased} weak learner to capture linguistic biases in IR datasets and then trains a robust model by regularizing and refining its predictions using the biased weak learner. This approach effectively prevents the robust model from overfitting to specific linguistic patterns in data. We propose four approaches for developing linguistically-biased models. Extensive experiments on several datasets show that our method reduces performance disparities across linguistically simple and complex queries, while improving overall retrieval performance.\looseness-1
\end{abstract}

\section{Introduction}

Neural ranking models have been extensively used in information retrieval and question answering tasks~\citep{10.1145/3397271.3401204,zhao-etal-2021-sparta,khattab2020colbert,karpukhin2020dense,xiong2021approximate,hofstatter2021efficiently}. 
We demonstrate that these models can show strong linguistic biases, where the retrieval performance is biased with respect to the ``linguistic complexity'' of queries, quantified by the variability and sophistication in productive vocabulary and grammatical structures in queries using existing tools~\citep{lu2010automatic,lu2012relationship,lee-etal-2021-pushing,lee-lee-2023-lftk}.\footnote{We consider lexical and syntactic linguistic complexity indicators in this study. Details of these indicators are provided in Appendix~\ref{sec:lc}, 
Table~\ref{tab:ling_ind}.}

\begin{figure}[t]
  \centering
  \includegraphics[width=0.45\textwidth]{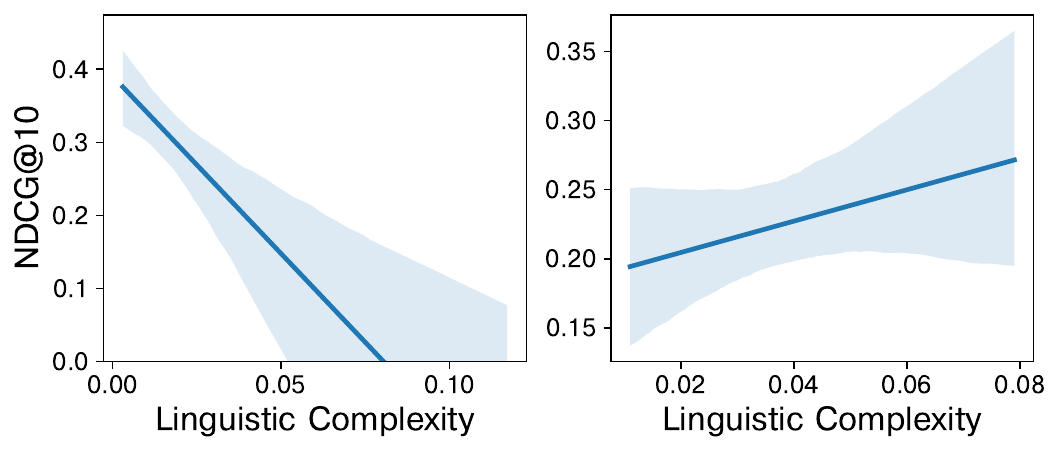}
  \caption{NDCG@10 of BM25 on the test set of NFCorpus~\citep{nfcorpus} (left) decreases and on the test set of FIQA~\citep{fiqa} (right) increases as the average linguistic complexity~\citep{lu2010automatic,lu2012relationship} of queries increase. 
  Specifically, we observe a significant drop in NDCG@10, from 0.4 to 0, and a significant increase in NDCG@10, from 0.2 to 0.3. The result shows that BM25 is significantly biased toward linguistically easy and hard examples on different datasets.}
  \vspace{-10pt}
  \label{fig:example}
\end{figure}

Figure~\ref{fig:example} shows that the average linguistic complexity of the test queries in the NFCorpus~\citep{nfcorpus} and FIQA~\citep{fiqa} datasets varies significantly, where the NDCG@10 performance of the BM25 model significantly decreases on NFCorpus and improves on FIQA as the linguistic complexity of queries increase. 
This performance disparity across queries of different linguistic complexity 
leads to the focus of this paper and the following research question:
\emph{can we debias IR models to achieve equitable performance across queries of varying linguistic complexity?}

\begin{figure*}[t]
  \centering
  \includegraphics[width=0.97\textwidth]{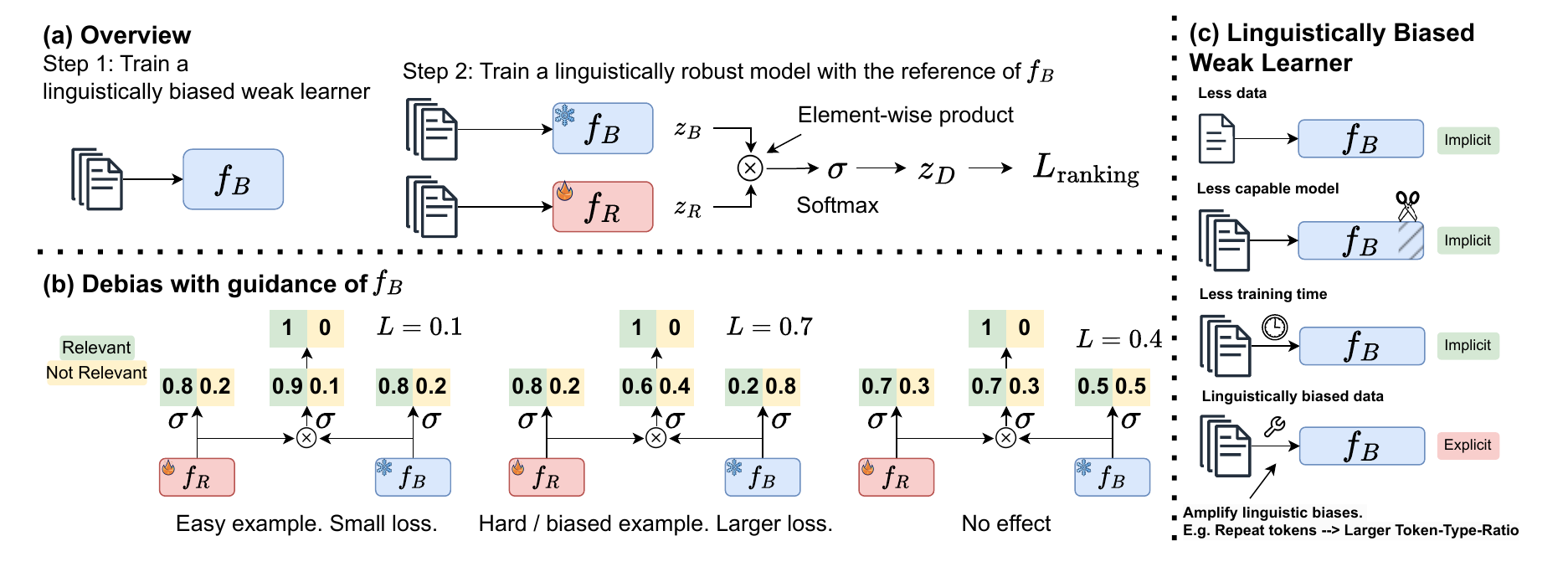}
  \caption{Architecture of \method for mitigating linguistic biases in IR models. (a) Training process: first, a linguistically biased IR model $f_B$ is trained. Then, we freeze the parameters of $f_B$ to train a target, linguistically robust IR model $f_R$ by taking the product of logits of $f_B$ and $f_R$. The biased weak learner regularizes the ranking loss of $f_R$ using its learned linguistic biases. 
  (b): Examples showing that the ensemble approach effectively moderates prediction probabilities to avoid learning biases associated with high confidence or moving too heavily toward the biased weak learner.
  (c): Strategies for developing linguistically biased weak learners. 
  }
  \label{fig:model}
\end{figure*}

Inspired by previous debiasing works in natural language processing~\citep{utama-etal-2020-towards,ghaddar-etal-2021-end,sanh2021learning,meissner-etal-2022-debiasing}, we introduce a new approach, named \method, to mitigate linguistic biases in IR models. \method is a {\em weak learner} framework; it first trains a linguistically-biased weak learner to explicitly capture linguistic biases in a dataset. This linguistically-biased weak learner is then used as a reference to inform and regularize the training of a desired (robust) IR model. It encourages the IR model to focus less on biased patterns and more on the underlying relevance signals. This is achieved by using the biased weak learner's predictions as indicators of bias intensity in inputs, and adjusting the IR model's predictions accordingly. 

\newpage
\method does not require linguistic biases to be explicitly described for the model, and reduces the risk of overfitting to specific types of biases. Specifically, we investigate several strategies to develop a linguistically-biased weak learner: 
training the model using \textbf{linguistically biased data} to directly introduce and reinforce specific linguistic patterns,
using a \textbf{weaker model} with fewer parameters or a simpler architecture to reduce models ability to generalize across inputs with various linguistic complexity,  
\textbf{shortening the training time} to prevent the model from capturing the diversity and depth of linguistic features in the data, and
training on a \textbf{limited data} to emphasize the linguistic features present in a specific subset of data.
Through these strategies, we aim to develop a model that effectively captures linguistic biases for developing linguistically robust IR models. 

Our contribution are 
(a): illustrating that the performance of current IR models vary based on the linguistic complexity of input queries, 
(b): a novel approach that trains a linguistically robust IR model with the help of a linguistically biased IR model to mitigate such biases, and
(c): four approaches to obtain linguistically biased weak learners, all effective in mitigating biases in IR models.

\section{\method}

\paragraph{Linguistic Complexity:}
measures sophistication in productive vocabulary and grammatical structures in textual content, spanning lexical, syntactic, and discourse dimensions. In this work, we adopt existing linguistic complexity measurements (lexical complexity~\citep{lu2012relationship} and syntactic complexity~\citep{lu2010automatic}) to measure the linguistic complexity of queries in IR datasets implemented by existing tools~\citep{lu2010automatic,lu2012relationship,lee-etal-2021-pushing,lee-lee-2023-lftk}. Specifically, given a query $q$, a linguistic complexity score is computed by averaging scores of various linguistic complexity metrics, which includes measures such as verb sophistication and the number of T-units. The detailed list of linguistic complexity is shown in Appendix~\ref{sec:lc} Table~\ref{tab:ling_ind}.
We column normalize linguistic complexity scores before computing average linguistic complexity for each query.


\paragraph{Overview:}
\method mitigates linguistic biases in an IR model using a linguistically-biased weak learner, $f_B$. The process begins with training $f_B$ to learn linguistic biases present in a dataset. Then, a linguistically robust model, $f_R$, is trained based on the confidence of $f_B$ (which approximates the intensity of linguistic biases in input) and the prediction accuracy of $f_R$. 
This approach has two purposes: 
firstly, $f_B$ guides $f_R$ to improve its robustness by learning from the identified biases of $f_B$. Secondly, $f_B$ can adjust the weights of training examples by prioritizing those that $f_R$ fails to predict, which effectively refines the training focus of $f_R$ toward more challenging examples.

\paragraph{Bi-Encoder Architecture:}
We consider a standard bi-encoder architecture with a query encoder $f_q$ and a document encoder $f_d$~\citep{khattab2020colbert,karpukhin2020dense,xiong2021approximate,hofstatter2021efficiently}. Given the $i$-th batch $\mathcal{B}_i = \{ q_i, d^+_i, d^-_{i, 1}, \dots, d^-_{i, n} \}$, where $q_i$ denotes the query, $d^+_i$ denotes a relevant document, and $d^-_{i, j}, \forall j$ denote irrelevant documents, we encode them into embeddings $h_{q_i}, h_{d^+_i}, h_{d^-_i}$, and optimize the standard contrastive loss:
\begin{equation}
    L = - \mathrm{log} \frac{e^{\mathrm{sim}(h_q, h_{d^+})}} {e^{\mathrm{sim}(h_q, h_{d^+})} + \sum_{j=1}^{n}e^{\mathrm{sim}(h_q, h_{d^-_{j}})}}
\end{equation}


\subsection{Debiasing with Biased Weak Learner}
We first train a linguistically biased weak learner $f_B$ using the bi-encoder architecture to model dataset biases. After training, we freeze $f_B$'s parameters and use it to train $f_R$. 
Given an input example $x_i = (q_i, d_i)$, we first obtain the logits
from the linguistically-biased weak learner $f_B$ and the target linguistically robust model $f_R$:
\begin{equation}\label{eq:zb_zr}
    z_B = f_B(x_i), \quad z_R = f_R(x_i).
\end{equation}

As Figure~\ref{fig:model}(a) shows, to integrate the knowledge from the linguistically biased weak learner into the training of the target IR model $f_R$, we compute the element-wise product of the two probabilities and normalize it with a softmax function, or more conveniently element-wise addition in log spcae: 
\begin{equation}
    \log(z_D) = \sigma \big(\alpha \log(z_B) + \log(z_R)\big),
\end{equation} 
where 
$\alpha \in [0, 1]$ is a scaling factor that controls the strength of the effect of the biases detected by $f_B$ on the final output of $f_R$.
This adjusted probability $z_D$ is the debiased probability (see the rationale below), which is then used to compute a standard ranking loss, where $f_B$ remains frozen and only the parameters of $f_R$ are updated. This approach encourages $f_R$ to adopt a less linguistically biased stance under the guidance of $f_B$. 

We note that the effect of element-wise product can be interpreted from two perspectives:
(a): dynamic curriculum: here the importance of training samples within a batch are adaptively re-weighted based on the confidence of $f_B$'s prediction; and 
(b): regularization function: here $f_B$ act as regularizer by constraining $f_R$ to avoid excessive confidence in its predictions, particularly for easy samples that it already predicts correctly. Consequently, $f_R$ does not overfit to specific biased patterns within the dataset. Therefore $f_B$ acts as both a guide and guard to make $f_R$ a more robust model against linguistic bias.
%

This approach effectively refines the training of $f_R$ using the weak learner $f_B$. Figure~\ref{fig:model}(b) provides several examples of the functionality of $f_B$. 
In case (1), when $f_B$ confidently makes a correct prediction, $f_R$ is adjusted to increase its confident in the correct label, as the input is likely an easy example. This lowers the loss (compared to $f_R$'s actual loss), reduces the weight of the example in training of $f_R$, and effectively minimizes the risk of learning biases from the example by $f_R$. 
In case (2), when $f_B$ confidently makes a wrong prediction, it indicates that the input sample likely contains biases that mislead $f_B$. Here, $f_R$'s confidence is adjusted to learn from the example by generating a larger than original loss, which encourages the model to adapt to these hard samples.


\subsection{Strategies for Developing Biased Learners}
Previous findings show that a ``weak'' model learns and relies on superficial patterns for making predictions~\citep{utama-etal-2020-towards,ghaddar-etal-2021-end,sanh2021learning,meissner-etal-2022-debiasing}. 
We introduce four approaches to obtain a linguistically-biased weak learner ($f_B$) from both model and data perspectives. 
\begin{itemize}
    \itemsep0pt
    \item First, we obtain a biased weak learner by \textbf{repeating linguistic constructs}, such as noun phrases, in queries. This approach makes the model more sensitive to complex linguistic structures by amplifying them in queries without changing the semantics. 

    \item Second, we train a \textbf{weaker model} with limited capacity to learn complex patterns, making it weaker in terms of predictive power but useful for exposing biases. This weaker model can be either a completely separate model (e.g. TinyBERT~\citep{turc2019well}) or a subset of $f_R$~\citep{cheng-amiri-2024-fairflow}.
%

    \item Third, we use the same architecture as the target IR model, but train it with significantly \textbf{fewer iterations}, which results in an ``undercooked'' version that is weaker. 

    \item Finally, we train the model on \textbf{less data}, which reduces its ability to generalize and learn deeper patterns.
\end{itemize}

Each of these weak learners reveal different linguistic biases in data, and provide insights into the biases that $f_R$ needs to overcome. 
Appendix~\ref{comp_fb}, Figure~\ref{fig:fb_bias} shows that the above approaches indeed result in linguistically biased $f_B$s.

\section{Experiments}

\paragraph{Datasets} We use the \emph{test} sets of four IR datasets form BEIR benchmark~\cite{thakur2021beir}:
\begin{itemize}
\itemsep0pt
    \item \textbf{MS MARCO}~\citep{nguyen2016ms}, a passage retrieval dataset with 532k training samples and 43 test queries; 
    \item \textbf{NFCorpus}~\citep{nfcorpus}, a biomedical IR dataset with 110k training samples and 323 test queries,
    \item \textbf{FIQA-2018}~\citep{fiqa}, a question answering dataset with 14k training samples and 648 test queries, and 
    \item \textbf{SciFact}~\citep{wadden-etal-2020-fact}, a scientific fact checking dataset with 920 training samples and 300 test queries.

\end{itemize}


\paragraph{IR Models} We compare our approach to the following baselines:
\begin{itemize}
    \itemsep0pt
    \item \textbf{BM25}~\citep{robertson2009probabilistic,manning2009introduction}, which retrieves documents based on lexical similarity; 
    \item \textbf{DPR}~\citep{karpukhin2020dense}, a dense retrieval model that compute similarity in embedding space;
    \item \textbf{ColBERT}~\citep{khattab2020colbert}, which adopts a delayed and deep interaction of token embeddings of query and document; 
    \item \textbf{Multiview}~\citep{amiri-etal-2021-attentive}, a multiview IR approach with data fusion and attention strategies; 

    \item \textbf{RankT5}~\citep{rankt5}, the Seq2Seq model~\citep{raffel2023exploring};
    \item \textbf{KernelWhitening}~\citep{gao-etal-2022-kernel}, which learns sentence embeddings that disentangles causal and spurious features; and 
    \item \textbf{LC as Rev Weight}, which uses linguistic complexity to reversely weight the probability. 
\end{itemize}

\paragraph{Evaluation}
Following previous works~\citep{thakur2021beir,rankt5}, we use NDCG@10 as the evaluation metric. We report average ($\mu, \uparrow$), standard deviation ($\sigma, \downarrow$), and coefficient of variation ($c_v = \frac{\sigma}{\mu}, \downarrow$) of NDCG@10 across all test queries. In addition, we examine models' performance in terms of the linguistic complexity of test examples. A robust model should have high overall performance and low performance variation across the spectrum of linguistic complexity (e.g. easy, medium, hard). Due to the limited space, we only implement \method to DPR.

\section{Main Results}

\begin{figure*}[t]
  \centering
  \includegraphics[width=0.97\textwidth]{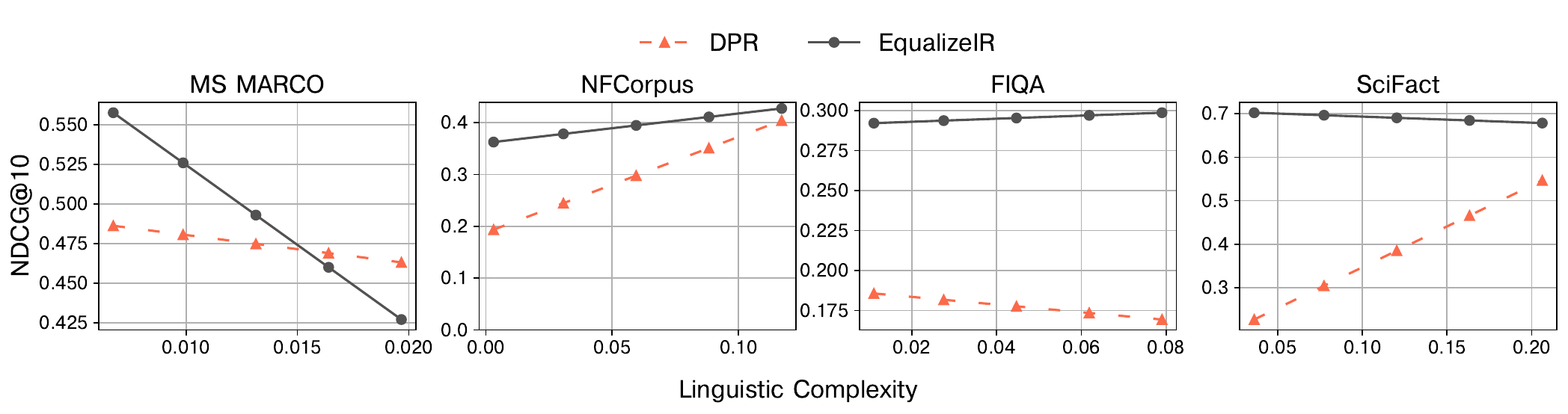}
  \caption{NDCG@10 of \method and DPR~\citep{karpukhin2020dense} as linguistic complexity of queries increase. Detailed performance of all baselines is shown in Figure~\ref{fig:ndcg} in Appendix~\ref{sec:app}.}
  \label{fig:ndcg_short}
\end{figure*}

\paragraph{Existing IR models are linguistically biased} 
Figure~\ref{fig:ndcg_short}~and~Table~\ref{tab:main_ave} show that existing IR models are linguistically biased with significant performance fluctuations as the linguistic complexity of query increases, resulting in a disparate performance across different levels of linguistic complexity. On average, BM25, DPR, ColBERT, RanKT5, and Multiview have varied performance across queries, with high standard deviation of 0.32, 0.32, 0.43, 0.25 and 0.26. These results highlight the need to mitigate linguistic biases in these models.  

\paragraph{\method increases average performance and reduces linguistic bias} \method outperforms BM25, DPR, ColBERT, RankT5, and Multiview by 0.03, 0.15, 0.15, 0.05, and 0.05 absolute points in average NDCG@10 respectively, while also showing smaller standard deviation in NDCG@10 across all test queries.
\method outperforms baselines in terms of $c_v$ (NDCG@10) by large margins of 0.30, 0.71, 1.19, 0.08, and 0.14 compared to BM25, DPR, ColBERT, RankT5, Multiview respectively.

\begin{table}[t]
\centering
\begin{tabular}{lrrr}
\toprule
Method & $\mu (\uparrow)$ & $\sigma (\downarrow)$ & $c_v (\downarrow)$ \\
\midrule
BM25      & \underline{0.44} & 0.32 & 0.82 \\
ColBERT   & 0.29 & 0.43 & 1.71 \\
DPR       & 0.29 & 0.32 & 1.23 \\
RankT5    & 0.42 & \underline{0.25} & \underline{0.64} \\
Multiview & 0.42 & 0.26 & 0.66 \\
KernelWhitening & 0.44 & 0.25 & 0.57 \\
LC as Rev Weight & 0.27 & 0.21 & 0.78 \\
\midrule
\method   & \textbf{0.47} & \textbf{0.22} & \textbf{0.52} \\
\bottomrule
\end{tabular}
\caption{Main results. $\mu$, $\sigma$, and $c_v$ denote average performance, standard deviation, and coefficient of variation across test queries. Best performance is in \textbf{bold} and second best is \underline{underlined}. The significance test is shown in Table~\ref{tab:sig}.}
\label{tab:main_ave}
\end{table}

\paragraph{Different IR models show different linguistic biases}
On NFCorpus, BM25 achieves 0.40 NDCG@10 on linguistically easy examples, while close to zero NDCG@10 on hard examples. Conversely, DPR perform poorly on linguistically easy examples and better on linguistically hard examples. This contrasting results can be attributed to the underlying architectures of the IR models, such as the text encoders and if late interaction is used, and the intrinsic characteristics of the datasets.


\paragraph{Comparison Between Different Biased Models}\label{comp_fb}
Figure~\ref{fig:fb_bias} shows that, as we hypothesized, all four types of weak learners encode substantial linguistic biases. Results in Appendix~\ref{sec:app} Table~\ref{tab:fb_fiqa}-\ref{tab:fb_scifact} show the comparison between different methods to obtain $f_B$. Overall, different $f_B$ training methods have similar overall performance and performance variation in terms of NDCG@10. We notice that that the ``weaker model'' and ``less data'' approaches consistently yield higher NDCG@10 performance, which may indicate that they better capture linguistic biases for $f_R$ to avoid. 
In contrast, the ``repeating linguistic constructs'' and ``fewer iterations'' strategies do not produce a good biased learner. This result could be attributed to the models potential overemphasis on specific linguistic features or lack of learning discriminative patterns from data, while overshadowing other aspects that may contribute to bias and resulting in a less effective bias detection. In addition, the ``weaker model'' and ``less data'' approaches may capture a broader type of biases, including implicit ones, which makes them more flexible and practical.
Using a less capable model as $f_B$ leads to the highest overall performance, smallest performance deviation and variation. Using less data has a slightly lower overall performance and higher performance deviation. This comparison highlights that different $f_B$s exhibit different linguistic biases and result in varying performances of $f_R$. \looseness-1

\section{Related Work}

\paragraph{Information Retrieval}
DPR~\citep{karpukhin2020dense} and ColBERT~\citep{khattab2020colbert} are earlier works of dense retrieval, where similarity is computed in high-dimensional embedding space. Although effective, \citet{10.1145/3626772.3657691} prove that operating in query-specific subspaces can improve the performance and efficiency of dense retrieval models. Recently, more attention has been paid to adapting Large Language Models (LLMs) to information retrieval~\citep{10.1145/3626772.3657819,10.1145/3626772.3657883,10.1145/3626772.3657917}.

\paragraph{Bias Mitigation}
\citet{li-etal-2022-debiasing} design an in-batch regularization technique to mitigate the biased performance across different subgroups. \citet{kim-etal-2024-discovering} propose to identify semantically relevant query-document pairs to explain why documents are retrieved, and discover that existing IR models show biased performances across different brand name. \citet{ziems-etal-2024-measuring} discover that IR models suffer from indexical bias, i.e. the bias resulted by the order of documents, and propose a new metric DUO to evaluate the amount of indexical bias an IR model has. Query performance prediction (QPP)~\citep{10.1007/978-3-031-56069-9_51} studies whether we can predict the IR quality by only looking at the query itself without additional information. On other tasks, prior works have discussed how biased models or weak learners can be applied to debiasing in vision~\citep{rubi}, natural language understanding~\citep{sanh2021learning,ghaddar-etal-2021-end,cheng-amiri-2024-fairflow}, and speech classification tasks~\citep{cheng24c_interspeech}.
\section{Conclusion}
We report that IR models are biased toward linguistic complexity of queries and introduce \method, a framework that trains a robust IR model by regularizing it with four types of linguistically-biased weak learners (by amplifying linguistic constructs in queries, using a weaker model with limited capacity, training with fewer iterations to create an underdeveloped model, and training on less data to restrict generalization), to achieve equitable performance across queries of varying linguistic complexity.\looseness-1

\section*{Limitations}
Existing definitions of linguistic complexity often have a narrow focus on specific linguistic features, which can result in challenges in comprehensive quantification of linguistic biases. For example, we did not consider linguistic biases related to discourse, pragmatics, morphology and semantics. 
In addition, our debiasing approach slightly increases complexity of training by requiring a trained biased model. 
Similar to other debiasing approaches, there's a risk of model overfitting to particular biases the model is trained to address, which may limit its adaptability to generalize to new or unseen biases. 
Finally, although our approach can be applied to any supervised IR model, we only applied it dense retrieval models, and its performance on other IR models remained underexplored. 

\section*{Broader Impact Statement}
We present an important issue in existing IR models: they show disparate and biased performance across queries with different levels of linguistic complexity--quantified by lexical and syntactic complexity. This can disproportionately disadvantage queries from users with specific writing style that result in particular types of linguistic complexity. 
It is important that future research and evaluation protocols in IR accounts for these biases and mitigate them.

\bibliography{anthology,reference}

\appendix
\newpage

\section{Addition Results}\label{sec:app}
\begin{figure*}[t]
  \centering
  \includegraphics[width=0.95\textwidth]{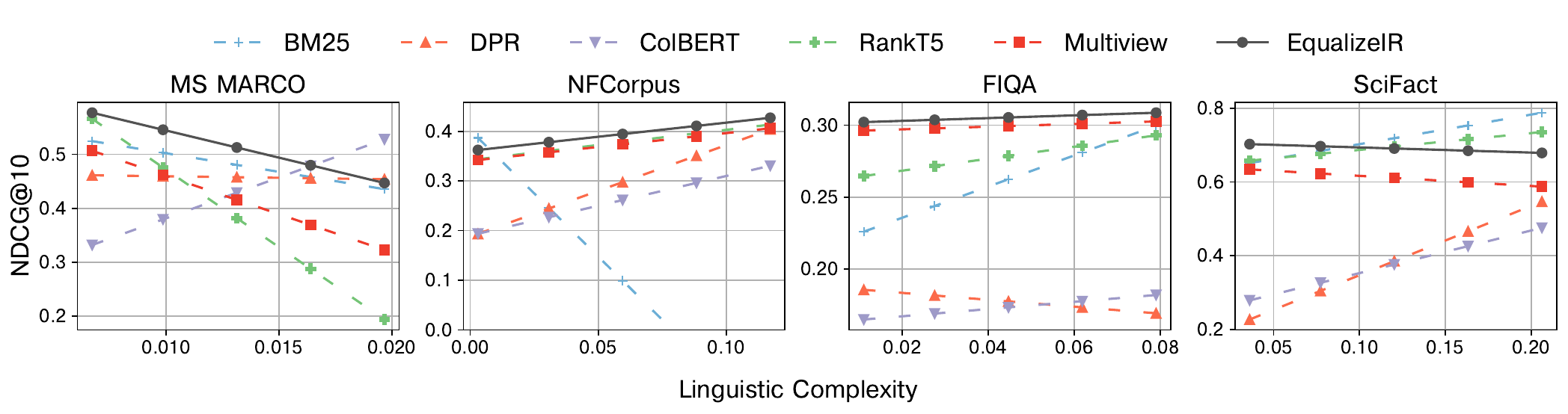}
  \caption{Performance in NDCG@10 as linguistic complexity of queries increase.}
  \label{fig:ndcg}
\end{figure*}

We present the performance with respect to linguistic complexity in Figure~\ref{fig:ndcg} and the performance on each dataset in Table~\ref{tab:main}. Overall, the results show that existing IR models are linguistically biased, showing significant performance fluctuations as the linguistic complexity of query changes. 
Table~\ref{tab:fb_fiqa}-\ref{tab:fb_scifact} compares the performances between different methods to obtain $f_B$.

\begin{figure*}[t]
  \centering
  \includegraphics[width=0.95\textwidth]{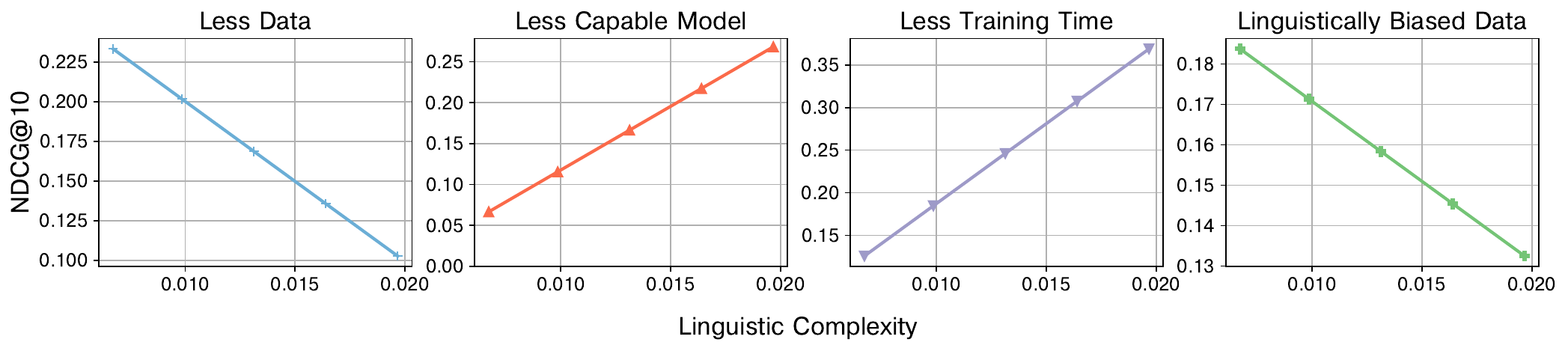}
  \caption{Performance of $f_B$ obtained by four different strategies, which are highly linguistically biased.}
  \label{fig:fb_bias}
\end{figure*}

\begin{table}[ht]
\tiny
\centering
\begin{tabular}{l|l|ccc}
\toprule
Data & Method & $\mu (\uparrow)$ & $\sigma (\downarrow)$ & $c_v (\downarrow)$ \\
\midrule
\multirow{6}{*}{FIQA} 
& BM25     & 0.25 & 0.32 & \underline{1.26} \\
& ColBERT  & 0.23 & 0.22 & 0.96 \\
& DPR      & 0.22 & \textbf{0.18} & 0.82 \\
& RankT5   & 0.26 & \underline{0.21} & 0.81 \\
& Multiview & 0.27 & 0.23 & 0.85 \\
& \method  & \textbf{0.29} & \underline{0.21} & \textbf{0.72} \\
\midrule
\multirow{6}{*}{MS MARCO} 
& BM25     & \textbf{0.48} & \underline{0.25} & \underline{0.53} \\
& ColBERT  & 0.44 & 0.38 & 0.88 \\
& DPR      & \underline{0.47} & 0.29 & 0.61 \\
& RankT5   & 0.43 & 0.33 & 0.77 \\
& Multiview & 0.42 & 0.35 & 0.83 \\
& \method  & \textbf{0.48} & \textbf{0.20} & \textbf{0.42} \\
\midrule
\multirow{6}{*}{NFCorpus} 
& BM25    & \underline{0.34} & 0.32 & 0.92 \\
& ColBERT & 0.28 & \underline{0.25} & 0.89 \\
& DPR     & 0.31 & 0.27 & \underline{0.87} \\
& RankT5  & 0.33 & 0.29 & 0.88 \\
& Multiview & 0.32 & 0.28 & 0.88 \\
& \method & \textbf{0.37} & \textbf{0.23} & \textbf{0.62} \\
\midrule
\multirow{6}{*}{SciFact} 
& BM25    & \underline{0.69} & \underline{0.39} & 0.56 \\
& ColBERT & 0.50 & 0.34 & 0.68 \\
& DPR     & 0.40 & 0.32 & 0.80 \\
& RankT5  & 0.68 & 0.33 & \underline{0.49} \\
& Multiview & 0.64 & 0.36 & 0.56 \\
& \method & \textbf{0.70} & \textbf{0.25} & \textbf{0.36} \\
\bottomrule
\end{tabular}
\caption{Main results. $\mu$, $\sigma$, and $c_v$ denote average performance, standard deviation, and coefficient of variation across all queries in each test set. Best performance is in \textbf{bold} and second best is \underline{underlined}.}
\label{tab:main}
\end{table}

\begin{table}[ht]
\small
\centering
\begin{tabular}{l|cccc}
\toprule
Data & MS MARCO	& NFCorpus & FIQA & SciFact \\
\midrule
BM25      & 2.8e-3	& 9.0e-4	& 2.2e-3	& 2.8e-2  \\ 
ColBERT   & 1.5e-3	& 2.0e-9	& 2.9e-12	& 1.1e-36 \\
DPR       & 2.9e-3	& 1.4e-13	& 3.3e-13	& 9.3e-38 \\
RankT5    & 1.1e-3	& 1.7e-4	& 9.1e-5	& 1.1e-2  \\
Multiview & 1.4e-5	& 6.0e-14	& 8.1e-14	& 1.4e-8  \\
\bottomrule
\end{tabular}
\caption{Significance test between EqualizeIR and baselines adjusted with bonferroni correction. Results show that EqualizeIR performs significantly better than baselines.}
\label{tab:sig}
\end{table}

\section{Linguistic Complexity}
\label{sec:lc}

Table~\ref{tab:ling_ind} presents the 45 linguistic complexity measurements in our study. For the full description of these metrics, see~\citep{lu2010automatic,lu2012relationship,lee-lee-2023-lftk}. We provide a brief description of a few indices as an example:  
\textbf{Type–Token Ratio, TTR} is the ratio of unique words in the text. \textbf{D-measure} is a modification to TTR that accounts for text length. \textbf{* Variation} indicates variations in lexical words such as nouns, verbs, adjectives, and adverbs. The \textbf{Mean Length of T-Units} is the average length of T-units in text. A T-unit is defined as a minimal terminable unit, essentially an independent clause and all its subordinate clauses. It provides insight into the syntactic complexity by measuring how elaborate the clauses are on average. 
%

\begin{table}[ht]
\centering
\resizebox{\linewidth}{!}{
\begin{tabular}{c|l|l}
\toprule
    \textbf{Type} & \textbf{Index Name} & \textbf{Notation} \\
    \midrule
    \multirow{13}{*}{\rotatebox{90}{\textbf{Syntactic}}} 
    & Mean length of clause & MLC \\
    & Mean length of sentence & MLS \\
    & Mean length of T-Unit & MLT \\
    & Sentence complexity ratio & C/S \\
    & T-unit complexity ratio & C/T \\
    & Complex T-unit proportion & CT/T \\
    & Dependent Clause proportion & DC/C \\
    & Dependent Clause to T-Unit ratio & DC/T \\
    & Sentence coordination ratio & T/S \\
    & Coordinate phrases to clause ratio & CP/C \\
    & Coordinate phrases to T-Unit ratio & CP/T \\
    & Complex nominals to clause ratio & CN/C \\
    & Complex nominals to T-unit ratio & CN/T \\
    & Verb phrases to T-unit ratio & VP/T \\
    \midrule
    \multirow{16}{*}{\rotatebox{90}{\textbf{Lexical}}} 
    & Type–Token Ratio TTR & T/N \\
    & Mean TTR of all 50-word segments & MSTTR–50 \\
    & Corrected TTR CTTR & T/$\sqrt{2N}$ \\
    & Root TTR RTTR & T/$\sqrt{N}$ \\
    & Bilogarithmic TTR & $\log(TTR)$ $\log(T)$ / $\log (N)$ \\
    & Uber Index Uber & $\log(2N)$ / $\log(N/T)$ \\
    & D Measure & D \\
    & Lexical Word Variation & LV Tlex/Nlex \\
    & Verb Variation-I & VV1 $T_{\mathrm{Verb}}$ / $N_{\mathrm{Verb}}$ \\
    & Squared VV1 & SVV1 Tv2 \\
    & Verb & $N_{\mathrm{Verb}}$ \\
    & Corrected VV1 & CVV1 $T_{\mathrm{Verb}}$ /$\sqrt{2Nverb}$ \\
    & Verb Variation-II & $T_{\mathrm{Verb}}$ /Nlex \\
    & Noun Variation & $T_{\mathrm{Noun}}$ / Nlex \\
    & Adjective Variation & AdjV $T_{\mathrm{Adj}}$ /Nlex \\
    & Adverb Variation & AdvV $T_{\mathrm{Adv}}$ /Nlex \\
    & Modifier Variation & ModV ($T_{\mathrm{Adj}}$ + $T_{\mathrm{Adv}}$ )/ Nlex \\
\bottomrule
\end{tabular}
}
\caption{Linguistic indices used in the study}
\label{tab:ling_ind}
\end{table}

\begin{table}[ht]
\small
\centering
\begin{tabular}{l|c|c|c}
\toprule
Dataset                     & $\mu (\uparrow)$ & $\sigma (\downarrow)$ & $c_v (\downarrow)$ \\
\midrule
Less data                   & \underline{0.27} & \underline{0.23} & \underline{0.85} \\
Less capable model          & \textbf{0.29} & \textbf{0.21} & \textbf{0.72} \\
Less trained                & \underline{0.27} & 0.24 & 0.89 \\
Linguistically biased data  & 0.26 & 0.26 & 1.01 \\
\bottomrule
\end{tabular}
\caption{Comparison of different strategies for developing linguistically biased models in terms of NDCG@10 on FIQA. Best performance is in \textbf{bold} and second best is \underline{underlined}.}
\label{tab:fb_fiqa}
\end{table}

\begin{table}[ht]
\small
\centering
\begin{tabular}{l|c|c|c}
\toprule
Dataset                     & $\mu (\uparrow)$ & $\sigma (\downarrow)$ & $c_v (\downarrow)$ \\
\midrule
Less data                   & \underline{0.44} & \underline{0.23} & \underline{0.52} \\
Less capable model          & \textbf{0.48} & \textbf{0.20} & \textbf{0.42} \\
Less trained                & 0.42 & 0.26 & 0.62 \\
Linguistically biased data  & 0.42 & 0.25 & 0.60 \\
\bottomrule
\end{tabular}
\caption{Comparison of different strategies for developing linguistically biased models in terms of NDCG@10 on MS MARCO. Best performance is in \textbf{bold} and second best is \underline{underlined}.}
\label{tab:fb_msmarco}
\end{table}

\begin{table}[ht]
\small
\centering
\begin{tabular}{l|c|c|c}
\toprule
Dataset                     & $\mu (\uparrow)$ & $\sigma (\downarrow)$ & $c_v (\downarrow)$ \\
\midrule
Less data                   & \underline{0.33} & \underline{0.27} & \underline{0.81} \\
Less capable model          & \textbf{0.37} & \textbf{0.23} & \textbf{0.62} \\
Less trained                & \underline{0.35} & 0.25 & 0.71 \\
Linguistically biased data  & 0.32 & 0.26 & 0.81 \\
\bottomrule
\end{tabular}
\caption{Comparison of different strategies for developing linguistically biased models in terms of NDCG@10 on NFCorpus. Best performance is in \textbf{bold} and second best is \underline{underlined}.}
\label{tab:fb_nfcorpus}
\end{table}

\begin{table}[ht]
\small
\centering
\begin{tabular}{l|c|c|c}
\toprule
Dataset                     & $\mu (\uparrow)$ & $\sigma (\downarrow)$ & $c_v (\downarrow)$ \\
\midrule
Less data                   & \underline{0.68} & \underline{0.33} & \underline{0.49} \\
Less capable model          & \textbf{0.70} & \textbf{0.25} & \textbf{0.36} \\
Less trained                & 0.67 & 0.35 & 0.52 \\
Linguistically biased data  & 0.61 & 0.30 & 0.49 \\
\bottomrule
\end{tabular}
\caption{Comparison of different strategies for developing linguistically biased models in terms of NDCG@10 on SciFact. Best performance is in \textbf{bold} and second best is \underline{underlined}.}
\label{tab:fb_scifact}
\end{table}

\section{Implementation Details}
We use PyTorch~\citep{pytorch} and BEIR~\citep{thakur2021beir} to implement our approach. For DPR and ColBERT, we use BERT-base~\citep{devlin-etal-2019-bert} as the encoders. For $f_B$ trained with less data, we randomly take 20\% of the original training data to train $f_B$. For $f_B$ trained with less capable model, we use BERT-Tiny~\citep{turc2019well} as the encoder. For $f_B$ trained with less time, we train it for 20\% of the original training time. 
All methods are trained with AdamW~\citep{loshchilov2018decoupled} optimizer with a learning rate of $1e-5$. We tune $\alpha$ on validation sets and find choosing $\alpha=0.1$ yields best performance consisitently across datasets.


\end{document}